\documentclass[final]{cvpr}

\usepackage{times}
\usepackage{epsfig}
\usepackage{graphicx}
\usepackage{amsmath}
\usepackage{amssymb}

\usepackage[pagebackref=true,breaklinks=true,colorlinks,bookmarks=false]{hyperref}

\def\cvprPaperID{1348} 
\def\confYear{CVPR 2021}

\usepackage[utf8]{inputenc} 
\usepackage[T1]{fontenc}    
\usepackage{hyperref}       
\usepackage{url}            
\usepackage{booktabs}       
\usepackage{amsfonts}       
\usepackage{nicefrac}       
\usepackage{microtype}      
\usepackage{xcolor}
\usepackage{graphicx}
\usepackage{diagbox}
\usepackage{multirow}
\usepackage{mmstyles}

\begin{document}
	
	\title{Bidirectional Projection Network for Cross Dimension Scene Understanding}
	
\author{Wenbo Hu$^{1,3}$\footnotemark[1]~ \quad Hengshuang Zhao$^{2}$\footnotemark[1]~ \quad Li Jiang$^{1}$ \quad Jiaya Jia$^{1}$ \quad Tien-Tsin Wong$^{1,3}$\footnotemark[2]~\\
	$^{1}$The Chinese University of Hong Kong \quad $^{2}$University of Oxford \\ $^{3}$Shenzhen Key Laboratory of Virtual Reality and Human Interaction Technology, SIAT, CAS\\ 
	{\tt\small \{wbhu, lijiang, leojia, ttwong\}@cse.cuhk.edu.hk} \quad {\tt\small hengshuang.zhao@eng.ox.ac.uk}
}


\maketitle  \renewcommand{\thefootnote}{\fnsymbol{footnote}}
\footnotetext[1]{Equal contribution.} \footnotetext[2]{Corresponding author.}

\pagestyle{empty}
\thispagestyle{empty}

\begin{abstract}
2D image representations are in regular grids and can be processed efficiently, whereas 3D point clouds are unordered and scattered in 3D space. 
The information inside these two visual domains is well complementary, e.g., 2D images have fine-grained texture while 3D point clouds contain plentiful geometry information.
However, most current visual recognition systems process them individually.
In this paper, we present a \emph{bidirectional projection network (BPNet)} for joint 2D and 3D reasoning in an end-to-end manner.
It contains 2D and 3D sub-networks with symmetric architectures, that are connected by our proposed \emph{bidirectional projection module (BPM)}.
Via the \emph{BPM}, complementary 2D and 3D information can interact with each other in multiple architectural levels, such that advantages in these two visual domains can be combined for better scene recognition.
Extensive quantitative and qualitative experimental evaluations show that joint reasoning over 2D and 3D visual domains can benefit both 2D and 3D scene understanding simultaneously.
Our \emph{BPNet} achieves top performance on the ScanNetV2 benchmark for both 2D and 3D semantic segmentation.
Code is available at \url{https://github.com/wbhu/BPNet}.

\end{abstract}


\vspace{-4mm}
\section{Introduction}
\label{sec:introduction}
Scene understanding is a fundamental while challenging problem in computer vision. Multiple sensors are used to capture the scene information. 
The 2D camera is the most common sensor in our daily life. It projects the 3D space to image planes with plenty of fine-grained textures captured. 2D images with pixels densely arranged in regular grids can be processed efficiently with deep convolutional neural networks. We have witnessed remarkable improvements on 2D visual reasoning, \eg, image classification~\cite{krizhevsky2012imagenet,szegedy2015going,simonyan2015very,he2016deep} and semantic segmentation~\cite{long2015fully,chen2015semantic,yu2016multi,zhao2017pyramid}. 
3D sensors, on the other hand, can provide important geometry information of the scene. 3D data is usually represented in points that are unordered and irregularly scattered in 3D space. Conventional convolution that relies on ordered grids can not be directly adapted to 3D data.
Hence, several tailor-made neural networks~\cite{qi2017pointnet,li2018pointcnn,sscn2018} have been proposed for 3D scene recognition and understanding.

We observe that the information inside 2D and 3D data is well complementary. 2D images provide detailed texture and color information while 3D point clouds contain strong shape and geometry knowledge.  
Although the techniques for individual 2D and 3D reasoning are studied a lot in the literature, the exploration of combining both 2D and 3D data for recognition is very limited. 
Existing methods that utilize both 2D and 3D data for recognition mostly adopt the unidirectional scheme.
For example, to incorporate 3D information for 2D scene understanding, some methods either encode depth into geocentric inputs~\cite{gupta2014learning} or incorporate it into convolution operations~\cite{qi20173d,wang2018depth}. 
But depth map only contains limited geometry information as it is view-dependent. The occluded part under the viewpoint together with the global context is missing. 
In the other aspect, 3DMV~\cite{dai20183dmv} and MVPNet~\cite{jaritz2019multi} utilize 2D information to assist the 3D recognition by first extracting multi-view image features and then lifting them into 3D space for fusing with the 3D features. 
However, we argue that the unidirectional scheme cannot fully leverage the complementary information inside the 2D and 3D data, bidirectionally interacting and fusing 2D and 3D features can better combine the advantages of these two visual domains as evidenced by our experiments.

\begin{figure*}[!t] 
	\centering
	\includegraphics[width=0.98\linewidth]{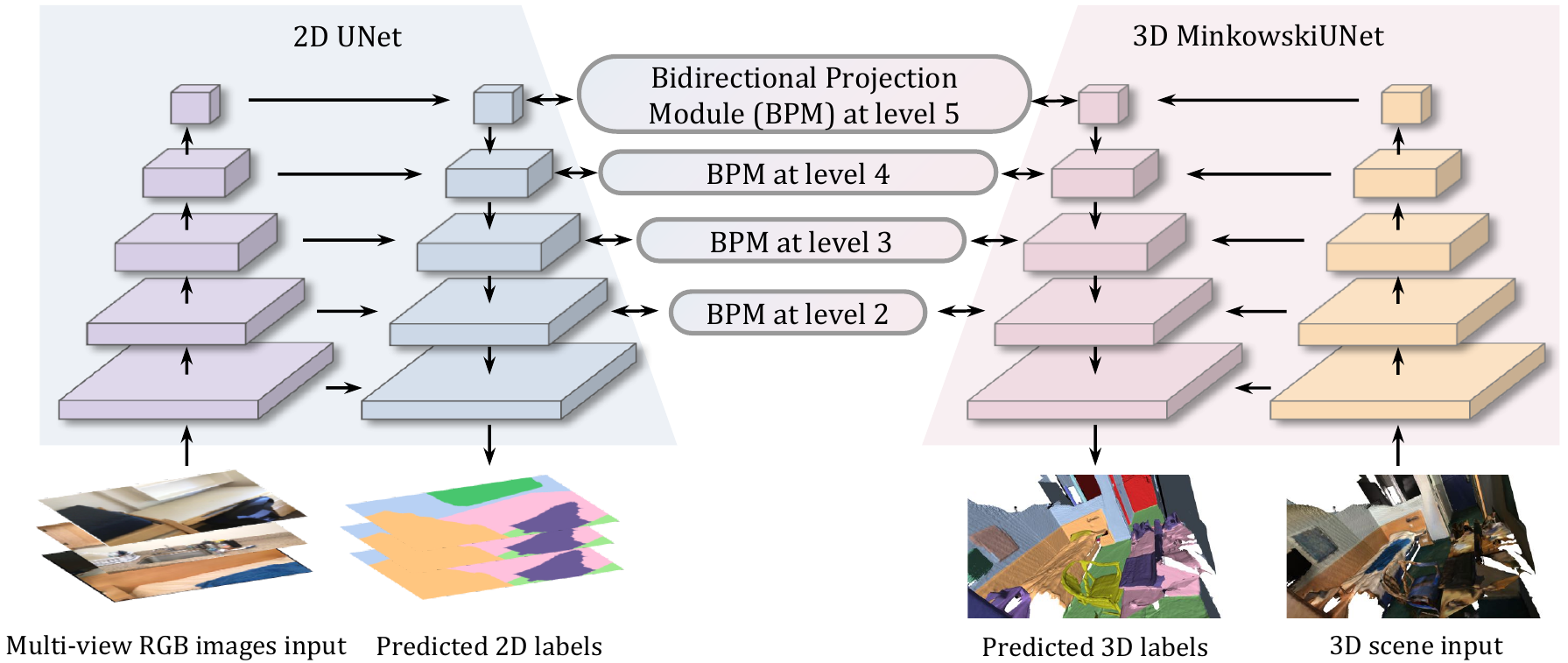}
	\caption{
		Overview of the Bidirectional Projection Network (BPNet). 
	It consists of two symmetric sub-networks, 2D UNet~\cite{ronneberger2015u} (in the left-hand side) and 3D MinkowskiUNet~\cite{choy20194d} (in the right-hand side).
	Both the 2D and 3D sub-networks are the U-shaped network with the same number of pyramid levels, but they are constructed based on conventional 2D convolution and 3D sparse convolution~\cite{sscn2018}, respectively.
	At multiple pyramid levels in the decoder stage, the features interact bidirectionally between 2D and 3D sub-networks via the proposed Bidirectional Projection Module (BPM). 
	Both 2D and 3D semantic labels will be predicted simultaneously by our BPNet. 
	}
		\vspace{-4mm}
	\label{fig:method}
\end{figure*}

In this paper,  we present a \emph{Bidirectional Projection Network (BPNet)} to enable information inside the 2D and 3D domains to flow bidirectionally at the network architectural level.
Such that the complementary information can be well combined for joint 2D and 3D scene understanding in an end-to-end manner.
Our method adopts two similar U-Net structures to process 2D and 3D data and introduces a \emph{Bidirectional Projection Module (BPM)} to bidirectionally fuse the multi-view 2D and 3D features. 
In this way, both 2D and 3D sides can benefit from each other. The overall framework is shown in Figure~\ref{fig:method}.
To be mentioned, we employ BPM at multiple pyramid levels, such that the features from 2D and 3D domains can be aggregated in a coarse-to-fine manner and the BPNet can harvest both low- and high-level complementary information. 
At each level, BPM builds the projection link matrix between 2D and 3D, and then transfers the features bidirectionally according to the link matrix, \ie, 2D features are projected into 3D space for boosting the recognition of 3D and vice the verse.

We evaluated our model on ScanNetV2~\cite{dai2017scannet} dataset for both 2D and 3D semantic segmentation tasks. BPNet achieves top performance on the benchmark in terms of mIoU and consistently outperforms the baseline with a single 2D/3D network. 
Also, the qualitative results show the effectiveness of combining 2D and 3D information. BPNet can distinguish objects without much shape difference (\eg, ``wall'' and ``picture'') in 3D segmentation. 
Meanwhile, 2D objects are better segmented with sharper boundaries thanks to the underlying geometric clues provided by 3D features.
Besides, we evaluated the generalization ability of BPNet for 2.5D data on the typical RGB-D dataset, NYUv2~\cite{Silberman:ECCV12}, and the results show BPNet performs favorably against the typical RGB-D and joint 2D-3D baselines.
We believe the proposed BPM is also advantageous to other tasks where both 2D and 3D resources are available, \eg, classification, detection, and instance segmentation.
Our contributions are summarized below.
	\vspace{-2mm}
\begin{itemize}
	\item We argue that 2D and 3D information is complementary for the understanding of each other and joint optimization over both 2D and 3D scenes is applicable and proved to be beneficial.
	\vspace{-2mm}
	\item We propose a Bidirectional Projection Module (BPM) that enables information interacting between the 2D and 3D representations. And such bidirectional projection operation can be adopted at multiple levels in the decoder stage.
	\vspace{-2mm}
	\item We present a novel framework named Bidirectional Projection Network (BPNet) for jointly reasoning over 2D and 3D scenes. Our method achieves top performance on the challenging large-scale ScanNetv2 benchmark for both 2D and 3D semantic segmentation tasks, which demonstrates its effectiveness.
\end{itemize}

\section{Related Work}
\label{sec:relatedWork}

\paragraph{2D semantic segmentation.}
\label{subsec:2d_semseg}
Semantic segmentation on 2D images has been significantly improved with deep neural networks. By replacing the last fully-connected layers in classification frameworks with convolution operations, FCN~\cite{long2015fully} can produce the per-pixel predictions. Several encoder-decoder architectures~\cite{ronneberger2015u,noh2015learning,badrinarayanan2017segnet} are proposed to utilize the information in low-level layers to help refine the segmentation outputs. 
The receptive field is vital for accurate scene understanding, thus, dilated convolutions~\cite{chen2015semantic,yu2016multi} are introduced to enlarge the receptive field. 
Contextual information is also proved to be effective as adopted in ~\cite{liu2015parsenet,ChenPKMY18,zhao2017pyramid,zhao2017icnet,yang2018denseaspp}. Meanwhile, attention models~\cite{zhao2018psanet,zhang2018context,huang2019ccnet} are also utilized in semantic segmentation for their abilities to capture long-range contextual relationships. 
The capability of these 2D-driven approaches is limited by the lack of geometric information, and their performance would be further boosted with other domain features like 3D geometric representations.

\vspace{-4mm}
\paragraph{3D semantic segmentation.}
\label{subsec:3d_semseg}
3D point clouds are irregularly scattered in 3D space, resulting in difficulties in 3d scene understanding, as convolution operation generally work on regular grids~\cite{zhao2019pointweb}. 
Point based frameworks mainly adopt Multilayer Perception (MLP)-style networks~\cite{qi2017pointnet,qi2017pointnet++}, including local region-feature enhancement~\cite{wang2019dynamic,xu2018spidercnn,zhao2019pointweb,hu2020randla}, kernel-based parametric convolution~\cite{wang2018deep,wu2019pointconv,thomas2019kpconv} and graph reasoning~\cite{wang2018local,wang2019graph,jiang2019hierarchical,li2019deepgcns,lei2020seggcn}. 
Apart from directly handling the irregular inputs, some methods transform the unordered point sets to ordered ones that can be processed by convolution operations, including voxelization, followed by 3D convolution approaches~\cite{maturana2015voxnet,song2017semantic,tchapmi2017segcloud,han2020occuseg} and the efficient generations~\cite{riegler2017octnet,sscn2018,choy20194d}. 
Multi-view mechanisms~\cite{su15mvcnn,su20153dassisted,qi2016volumetric,akundu2020virtualMVFusion} are also widely adopted where 3D point clouds are projected into 2D images via different camera views. 
Besides, convolution on meshes~\cite{schult2020dualconvmesh}, fused point-voxel representation~\cite{zhang12356deep}, occupancy-aware design~\cite{han2020occuseg}, and multi-task learning~\cite{hu2020jsenet} are explored to improve the 3D scene understanding performance.
However, the 3D source lacks detailed textures and color information, resulting in limited performance. And it would be improved with the help of 2D representations.

\vspace{-4mm}
\paragraph{Recognition with combined 2D-3D data.}
\label{subsec:2d-3d_semseg}

There exist several approaches that unidirectionally fuse 2D and 3D information for improving the recognition performance in the 2D or 3D domain. 
To incorporate 3D information for 2D scene understanding, Gupta \etal.~\cite{gupta2014learning} encode depth as HHA images to provide the geocentric information,
3DGNN~\cite{qi20173d} utilizes 3D graph neural networks for gradually refining the semantic representations and Depth-aware CNN~\cite{wang2018depth} presents a depth-aware convolution to cooperate depth information. 
On the other hand, some methods are dedicated to introducing 2D information into 3D scene understanding,
\eg, back projecting multi-view 2D features to 3D volumes~\cite{dai20183dmv}, or point clouds~\cite{jaritz2019multi}, then aggregating them with the original 3D features;
and extracting features from texture patches~\cite{huang2019texturenet} for the 3D semantic segmentation using surface parameterization.
Different from them, we conduct cross-dimension reasoning which enables bidirectional 2D and 3D feature interaction, such that both 2D and 3D semantic segmentation can benefit from each other.

Besides, 
AutoContext~\cite{gadde2017efficient} employs a decision tree to fuse up the hand-crafted features from images and point clouds for facade segmentation, 
xMUDA~\cite{jaritz2020xmuda} presents cross-modal unsupervised domain adaption to improve performance on the image or point cloud domain.
SurfaceNet~\cite{ji2017surfacenet} encodes camera parameters and images in a 3D voxel representation and adopts 3D CNN on it for multiview stereopsis.
Moreover, SPLATNet~\cite{su2018splatnet} presents to represent point clouds in a high-dimensional lattice that enables fusing image and point features in the high-dimensional lattice for facade and 3D part segmentation. 
However, this 2D-3D fusion relies on the tailor-made bilateral convolution on the high-dimensional lattice. 
Different from them, our method can be directly applied to the conventional 2D and 3D CNNs for interacting 2D and 3D features bidirectionally in multi-levels.

\section{Methodology}
\label{sec:method}
\vspace{-2mm}

The goal of our method is to jointly predict the 2D and 3D semantics on real-world data, given 3D scenes and 2D image sequences along with corresponding camera matrix. In this section, we first present our Bidirectional Projection Module (BPM) (Figure~\ref{fig:bpm}) that enables cross-dimension feature interaction between 2D and 3D feature representations. Then we introduce the design details of our Bidirectional Projection Network (BPNet) (Figure~\ref{fig:method}). Finally, we give our implementation details of the proposed method.

\vspace{-1mm}
\subsection{Bidirectional Projection Module}
\label{subsec:bpm}
\vspace{-2mm}

\begin{figure*}[!t] 
	\centering
	\includegraphics[width=0.97\linewidth]{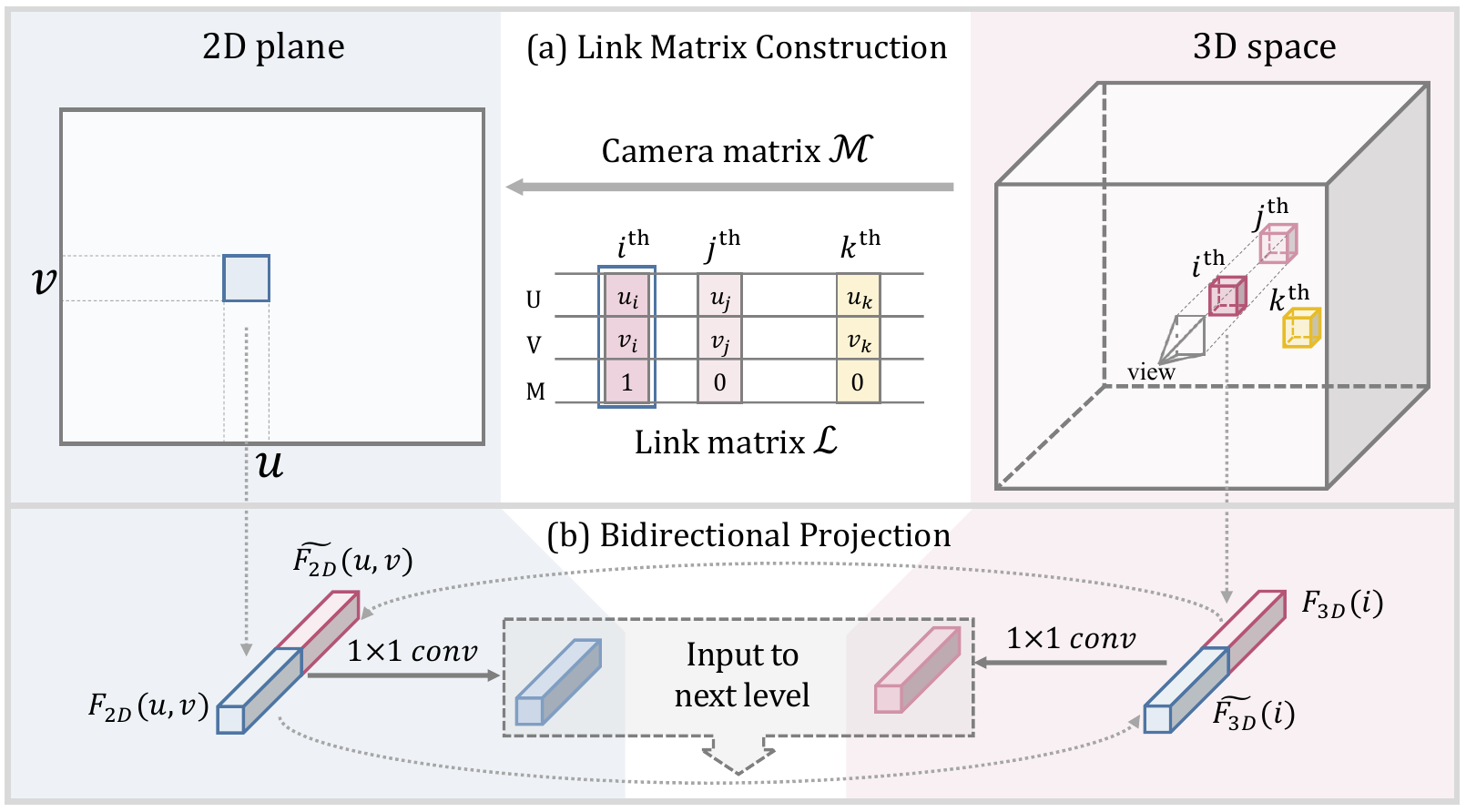}
	\caption{
		Bidirectional Projection Module (BPM). The link matrix construction procedure between one view and the 3D scene is illustrated in (a), and the bidirectional projection procedure is shown in (b).
	$\mathcal{M}$ is the camera matrix; the $i^\text{th}$, $j^\text{th}$ and $k^\text{th}$ voxels are typical examples of three kinds of voxels: voxels that have corresponding pixels under this view, voxels that are occluded by others, and voxels that are out of the view frustum. $F_\text{2D}$ and $F_\text{3D}$ are features in 2D and 3D spaces, respectively. $\widetilde{F_\text{2D}}$ and $\widetilde{F_\text{3D}}$ are the 2D and 3D bidirectionally projected features from the other domain, respectively.
	}
	\vspace{-5mm}
	\label{fig:bpm}
\end{figure*} 

The Bidirectional Projection Module (BPM), as shown in Figure~\ref{fig:bpm}, is designed to construct skip connections between 2D and 3D sub-networks at the same decoder level for bidirectionally interacting features in 2D and 3D domains, such that the strengths of these two visual domains can be integrated to boost both 2D and 3D scene understanding.
Specifically, given the 3D scene, and 2D image together with camera matrix $\mathcal{M}$, we first construct the link matrix $\mathcal{L}$ between voxels and pixels according to the perspective projection from 3D to 2D space.
Then, at multiple levels in the decoding stage, we not only project the 3D features $F_{3D}$ to 2D space but also back-project 2D features $F_{2D}$ to 3D space according to the constructed link matrix.
And finally, we concatenate the projected features with the original features followed by a $1\times1$ convolution to fuse them. We then feed the fused features into the next levels.
The detailed procedures in BPM are discussed in the following paragraphs.

\vspace{-4mm}
\paragraph{Link matrix construction.}
The camera imaging process can be seen as a perspective projection from 3D voxels to 2D pixels.
Mathematically, we can formulate it as:
\begin{equation}
	\vspace{-1mm}
	[u_i,v_i,1]^\text{T} = \mathcal{M}[x_i,y_i,z_i ,1]^\text{T},
	\label{eq:camera}
	\vspace{-1mm}
\end{equation}

where $[x_i,y_i,z_i ,1]^\text{T}$ and $[u_i,v_i,1]^\text{T}$ are the homogeneous 3D coordinates of the $i^{th}$ voxel and its projected 2D homogeneous coordinates, respectively; and $\mathcal{M}$ is the matrix that is the product of the intrinsic camera calibration matrix and the extrinsic camera pose matrix.

Based on it, we can create a link matrix $\mathcal{L}$ to store the corresponding 2D pixel coordinate $[u,v]^\text{T}$ for each voxel $[x,y,z]^\text{T}$, as shown in part (a) of Figure~\ref{fig:bpm}.
But some voxels may have no corresponding pixels, since the image under a certain viewpoint only captures part of the whole 3D scene, like the $k^\text{th}$ voxel in part (a) of Figure~\ref{fig:bpm}.
Therefore, we further add a mask $m$ with binary values to indicate whether this 3D voxel has a corresponding pixel.
Due to the occlusion, another problem is the hidden surfaces (or voxels) that can be projected to certain pixels but have no relations with the pixels at all, like the $j^\text{th}$ voxel in part (a) of Figure~\ref{fig:bpm} is occluded by the $i^\text{th}$ voxel.
In computer graphics, Z-Buffer algorithm~\cite{catmull1974subdivision} is commonly used to handle the hidden surface removal problem.
It is also applicable to our case, but the depth map is available here, thus, we can efficiently determine whether a voxel is hidden or not by comparing the depth value and the $z$ coordinate of the projected voxel.
Compared with the Z-Buffer algorithm, doing so can speed up the computing, as we compare the projected $z$ coordinate only once for each voxel.

As a result, we construct link matrix $\mathcal{L}$ as an $N \times 3$ matrix, $[U, V, M]$, where $N$ is the number of voxels in the 3D scene; $U$ and $V$ are the perspectively projected 2D coordinate according to Equation~\ref{eq:camera}; and $M$ is the binary mask to indicate whether this row of the link matrix is valid or not.
Each row of the link matrix $\mathcal{L}$ can be represented as:
\begin{equation}
	\mathcal{L}_i = \left\{
	\begin{array}{ll}
		& \text{if}      \qquad    U_{min}\leq u_i \leq U_{max}             \\
		{[u_i,v_i,1]}, & \text{and}  \quad\,      V_{min}\leq v_i \leq V_{max}             \\
		& \text{and}      \quad \,  |d(u_i,v_i) - z_i^{\prime}| \leq \delta \\
		&                                                                   \\
		{[u_i,v_i,0]}, & \text{otherwise}
	\end{array}
	\right.
		\vspace{-1mm}
\end{equation}
where $i \in [1,N]$ is the index of the voxels; $[u_i,v_i]^T$ is the 2D coordinate; 1 or 0 is the binary mask value; $U_{min},U_{max},V_{min}\,and\,V_{max}$ are the boundaries of the view frustum; $d(\cdot)$ is the mapping from pixel coordinates to the depth; $z_i^{\prime}$ is the projected $z$ coordinate of the voxel; and $\delta>0$ is the threshold for depth matching (it is set to be the voxel size in our implementation.) 
Note that, this link matrix is bidirectional, it can be used both to transform 2D points to 3D points and vice versa.

\vspace{-5mm}
\paragraph{Bidirectional projection.}
The link matrix is constructed for the original input 3D voxels and 2D images, while the 2D and 3D features have different spatial sizes compared with the original input data.
To produce the link matrix for 2D and 3D features in certain decoder levels, we linearly remap the $[u,v]$ coordinates according to the down-sampling ratio to fit for the spatial size of features in the link vectors.

Having the remapped links for each level of features, we can transform the features bidirectionally between 2D and 3D domains in multiple levels, as shown in part (b) of Figure~\ref{fig:bpm}.
At one of the decoder levels, we have the 2D feature $F_\text{2D}$ with the shape of $H \times W\times C_\text{2D}$, and the 3D feature $F_\text{3D}$ with the shape of  $N \times C_\text{3D}$ , where $N$ is the number of voxels, $C_\text{2D}$ and $C_\text{3D}$ are the channel numbers of 2D and 3D, respectively.
Meanwhile, the constructed link matrix $\mathcal{L}$ has the shape of $N \times 3$, where ``$3$'' indicates the corresponding $u,v$ coordinate of 2D feature and validity mask $m$.
To project 3D features to 2D space, we can formulate it as:
\begin{equation}
	\widetilde{F_\text{2D}}(u_i,v_i) = \left \{
	\begin{array}{ll}
		{m_i \cdot F_\text{3D}(i)}, & \text{if}\; [u_i,v_i]  \in \mathcal{L}_{uv}            \\
		
		\\
		{\vzero}, & \text{otherwise}
	\end{array}
	\right.
		\vspace{-2mm}
\end{equation}
where $\widetilde{F_\text{2D}}$ is the projected 2D feature from 3D feature that has the shape of $H \times W \times C_\text{3D}$; $i \in [1,N]$ is the index of voxels or link matrix entries; $[u_i,v_i,m_i]$  is the $i^{\text{th}}$ entry of $\mathcal{L}$; and $\mathcal{L}_{uv}$ is the first two columns of the link matrix $\mathcal{L}$.
On the other hand, to back project 2D features to 3D space, we formulate it as:
	\vspace{-1mm}
\begin{equation}
	\widetilde{F_\text{3D}}(i) = m_i  \cdot  F_\text{2D}(u_i,v_i),
	\vspace{-2mm}
\end{equation}
where $\widetilde{F_\text{3D}}$ is the back-projected feature from 2D feature with the shape of $N \times C_\text{2D}$.
These procedures can be implemented efficiently by the fanny indexing in PyTorch~\cite{paszke2019pytorch} via our constructed link matrix.

\vspace{-5mm}
\paragraph{View fusion.}
The above discussion is based on a 3D scene with a single 2D view.
For a 3D scene with multiple 2D views, we can directly project 3D features to each view with the corresponding link matrix.
But to transform multi-view 2D features to 3D space, we need to fuse them after back-projecting them to 3D space.
%
%
Different from 3DMV~\cite{dai20183dmv} that simply aggregates multi-view features by max-pooling, we use two-layer sparse convolutions to learn the impact factors for each view at every point and then weighted sum them up by the learned impact factors as following:
	\vspace{-2mm}
\begin{equation}
	\widetilde{F_\text{3D}}^{\prime} = \sum_{r=1}^{R}w_r \cdot \widetilde{F_\text{3D}^r},
	\vspace{-2mm}
\end{equation}
where $R$ is the number of views and $w_r$ is a vector of weights that learned from all the back-projected features.

\subsection{Bidirectional Projection Network}
\vspace{-1mm}
Based on the proposed BPM, we give our full bidirectional projection network (BPNet) as illustrated in Figure~\ref{fig:method}. The 2D and 3D sub-networks in our BPNet are both U-shaped networks with residual blocks~\cite{he2016deep}, but based on conventional 2D convolution and sparse 3D convolution~\cite{sscn2018}, respectively. 
During the training phase, we randomly sample $n$ 2D views to maintain the diversity of 2D data, while during the testing phase, we divide the 2D frames into $n$ groups and select one central view from each group to reduce the overlap among 2D views.
In our implementation, $n$ is set to three and the ablation is to be discussed in Section~\ref{sec:experiment}.

We firstly voxelize the 3D point clouds into volumes, and then the voxels represented in the sparse tensor format are fed into the 3D MinkowskiUNet, while the multi-view 2D images are simultaneously fed into the 2D UNet. 
During the decoder stages, we bidirectionally interact the features between 2D and 3D sub-networks in multiple pyramid levels, \ie, P2, P3, P4, and P5 levels, via the proposed BPM, thus harvesting both the low- and high-level features in these two domains. And finally, both 2D and 3D semantic labels are predicted from 2D and 3D parts in our BPNet, respectively.

\subsection{Implementation}
\vspace{-1mm}
As our goal is simultaneously predicting the 2D and 3D semantic labels, the loss function $\mathbf{L}$ consists of two terms:
\begin{equation}
	\label{loss}
	\mathbf{L} = \mathbf{H}_\text{3d} + \lambda \cdot \mathbf{H}_\text{2d},
	\vspace{-2mm}
\end{equation}
where $\mathbf{H}_\text{3d}$ and $\mathbf{H}_\text{2d}$ are the cross entropy losses for 3D and 2D predictions, respectively; and $\lambda$ is a weight to balance the 2D and 3D losses, which is empirically set to $0.1$ in our experiments.

We implement our algorithm based on the PyTorch~\cite{paszke2019pytorch} platform and MinkowskiEngine~\cite{choy20194d}
sparse convolution library.
We train our BPNet on the ScanNetV2 dataset~\cite{dai2017scannet} for 100 epochs.
The 2D UNet part is initialized from the classification weights pretrained on ImageNet~\cite{deng2009imagenet}, while the 3D part is initialized from scratch.
We use SGD solver~\cite{bottou2010large} with a base learning rate of 0.01 and a mini-batch size of 16. And we employ a poly learning rate scheduler with the power set to $0.9$. Momentum and weight decay are set to $0.9$ and $0.0001$, respectively.


\begin{table*}[!t]
	\scriptsize
	\centering
	\renewcommand{\tabcolsep}{0.92pt}
	\renewcommand{\arraystretch}{1.3}
	\resizebox{1.0\linewidth}{!}{
	\begin{tabular}{l|c|cccccccccccccccccccc}
		\toprule[1pt] 
		Method                                   &  mIoU  &  bath  &  bed   & bkshf  &  cab   & chair  &  cntr  &  curt  &  desk  &  door  & floor  & other  &  pic   & fridge & shower &  sink  &  sofa  & table & toilet &  wall  & window \\ \hline
		PointNet++~\cite{qi2017pointnet++}       &  33.9  &  58.4  &  47.8  &  45.8  &  25.6  &  36.0  &  25.0  &  24.7  &  27.8  &  26.1  &  67.7  &  18.3  &  11.7  &  21.2  &  14.5  &  36.4  &  34.6  & 23.2  &  54.8  &  52.3  &  25.2  \\
		SPLATNet$^\dagger$~\cite{su2018splatnet}  &  39.3  &  47.2  &  51.1  &  60.6  &  31.1  &  65.6  &  24.5  &  40.5  &  32.8  &  19.7  &  92.7  &  22.7  &  00.0  &  00.1  &  24.9  &  27.1  &  51.0  & 38.3  &  59.3  &  69.9  &  26.7  \\
		3DMV$^\dagger$~\cite{dai20183dmv}        &  48.4  &  48.4  &  53.8  &  64.3  &  42.4  &  60.6  &  31.0  &  57.4  &  43.3  &  37.8  &  79.6  &  30.1  &  21.4  &  53.7  &  20.8  &  47.2  &  50.7  & 41.3  &  69.3  &  60.2  &  53.9  \\ 
		FAConv\cite{zhang2020fusion}             &  63.0  &  60.4  &  74.1  &  76.6  &  59.0  &  74.7  &  50.1  &  73.4  &  50.3  &  52.7  &  91.9  &  45.4  & {32.3} &  55.0  &  42.0  &  67.8  &  68.8  & 54.4  &  89.6  &  79.5  &  62.7  \\
		MCCNN~\cite{hermosilla2018monte}         &  63.3  & {86.6} &  73.1  &  77.1  &  57.6  &  80.9  &  41.0  &  68.4  &  49.7  &  49.1  &  94.9  &  46.6  &  10.5  &  58.1  &  64.6  &  62.0  &  68.0  & 54.2  &  81.7  &  79.5  &  61.8  \\
		FPConv~\cite{lin2020fpconv}              &  63.9  &  78.5  &  76.0  &  71.3  &  60.3  &  79.8  &  39.2  &  53.4  &  60.3  &  52.4  &  94.8  &  45.7  &  25.0  &  53.8  &  72.3  &  59.8  &  69.6  & 61.4  &  87.2  &  79.9  &  56.7  \\
		MVPNet$^\dagger$~\cite{jaritz2019multi}  &  64.1  &  83.1  &  71.5  &  67.1  &  59.0  &  78.1  &  39.4  &  67.9  &  64.2  &  55.3  &  93.7  &  46.2  &  25.6  &  64.9  &  40.6  &  62.6  &  69.1  & 66.6  &  87.7  &  79.2  &  60.8  \\
		DCM-Net~\cite{schult2020dualconvmesh}    &  65.8  &  77.8  &  70.2  &  80.6  &  61.9  &  81.3  &  46.8  &  69.3  &  49.4  &  52.4  &  94.1  &  44.9  &  29.8  &  51.0  &  82.1  &  67.5  &  72.7  & 56.8  &  82.6  &  80.3  &  63.7  \\
		PointConv~\cite{wu2019pointconv}         &  66.6  &  78.1  &  75.9  &  69.9  &  64.4  &  82.2  &  47.5  &  77.9  &  56.4  &  50.4  &  95.3  &  42.8  &  20.3  &  58.6  &  75.4  &  66.1  &  75.3  & 58.8  & {90.2} &  81.3  &  64.2  \\
		PointASNL~\cite{yan2020pointasnl}        &  66.6  &  70.3  &  78.1  &  75.1  &  65.5  &  83.0  &  47.1  &  76.9  &  47.4  &  53.7  &  95.1  &  47.5  &  27.9  &  63.5  &  69.8  &  67.5  &  75.1  & 55.3  &  81.6  &  80.6  &  70.3  \\
		KP-FCNN~\cite{thomas2019kpconv}          &  68.4  &  84.7  &  75.8  &  78.4  &  64.7  &  81.4  &  47.3  &  77.2  &  60.5  &  59.4  &  93.5  &  45.0  &  18.1  &  58.7  &  80.5  &  69.0  &  78.5  & 61.4  &  88.2  &  81.9  &  63.2  \\
		MinkowskiNet~\cite{choy20194d}           &  73.6  &  85.9  & {81.8} & {83.2} &  70.9  & {84.0} & {52.1} & {85.3} &  66.0  &  64.3  &  95.1  & {54.4} &  28.6  &  73.1  & {89.3} &  67.5  &  77.2  & 68.3  &  87.4  &  85.2  & {72.7} \\ 
		{BPNet} (Ours)$^\dagger$                 &  74.9  &  90.9  &  81.8  &  81.1  &  75.2  &  83.9  &  48.5  &  84.2  &  67.3  &  64.4  &  95.7  &  52.8  &  30.5  &  77.3  &  85.9  &  78.8  &  81.8  & 69.3  &  91.6  &  85.6  &  72.3  \\
		\bottomrule[1pt]
	\end{tabular}}
	\vspace*{0.5mm}
	\caption{Comparison with the typical streams of methods on ScanNetV2 3D Semantic label benchmark, including point cloud based, sparse convolution based, and joint 2D-3D-input (marked with $\dagger$) based methods.
}
	\label{tab:seg_3d}
	\vspace*{-2mm}
\end{table*}

\begin{figure*}[!t] 
    \centering
    \includegraphics[width=0.985\linewidth]{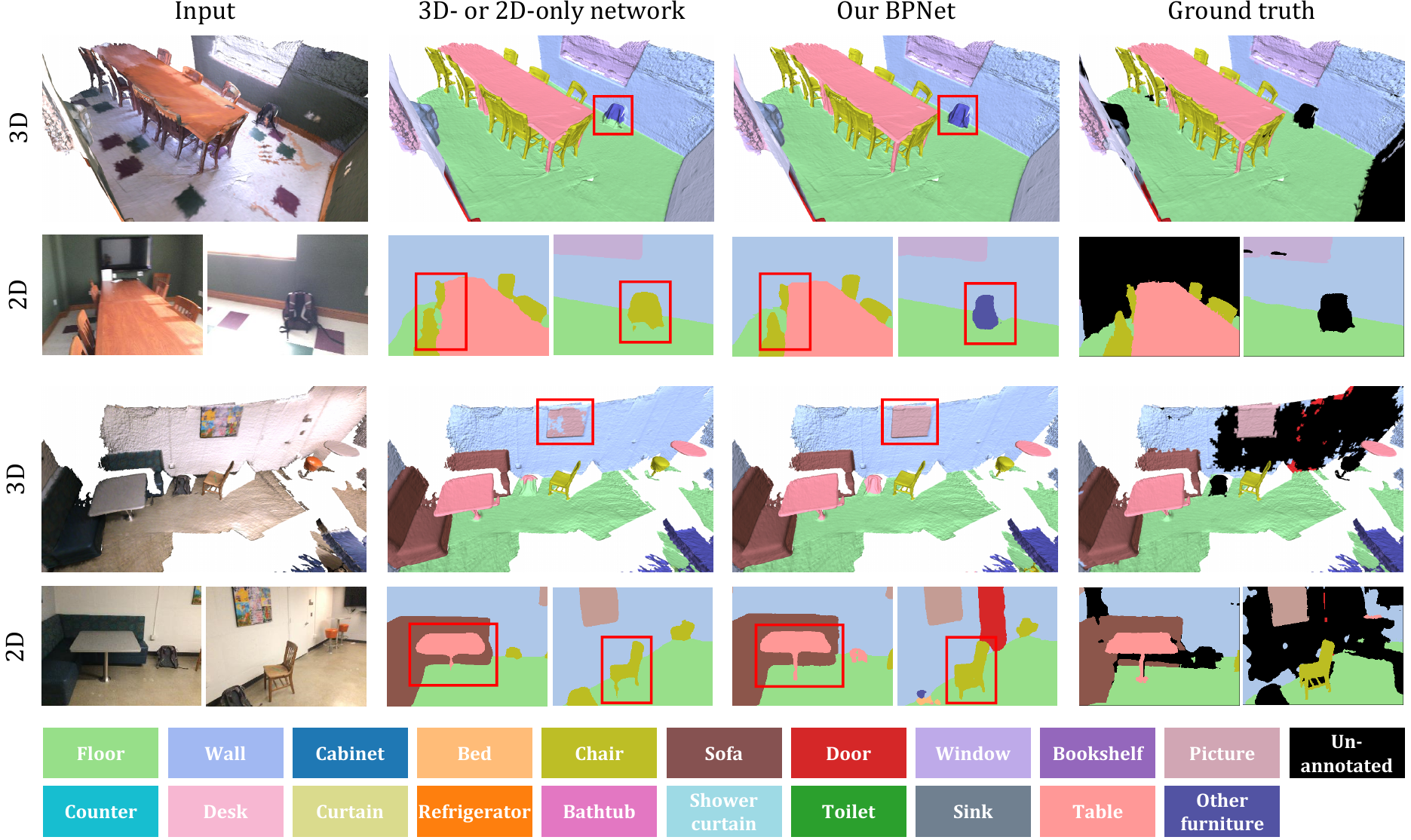}
    \caption{
        Qualitative 3D and 2D result examples of the 3D-only network, MinkowskiNet~\cite{choy20194d}, 2D-only network, UNet34, our BPNet, and the ground truths. Different semantics are labeled as corresponding colors as shown in the bottom color palette. We highlight the differences between the results of BPNet and others by red boxes.
    }
    \vspace*{-5mm}
    \label{fig:seg_3d}
\end{figure*} 

\vspace{-2mm}
\section{Experimental Evaluation}
\label{sec:experiment}

\begin{table*}[!t]
    \scriptsize
    \centering
    \renewcommand{\tabcolsep}{1.25pt}
    \renewcommand{\arraystretch}{1.3}
    
    \resizebox{1.0\linewidth}{!}{
        \begin{tabular}{l|c|cccccccccccccccccccc}
            \toprule[1pt] 
            Method                             &  mIoU  &  bath  &  bed   & bkshf  &  cab   & chair  &  cntr  &  curt  &  desk  &  door  & floor  & other  &  pic   & fridge & shower &  sink  &  sofa  & table  & toilet &  wall  & window \\ \hline
            PSPNet~\cite{zhao2017pyramid}      &  47.5  &  49.0  &  58.1  &  28.9  &  50.7  &  6.7   &  37.9  &  61.0  &  41.7  &  43.5  &  82.2  &  27.8  &  26.7  &  50.3  &  22.8  &  61.6  &  53.3  &  37.5  &  82.0  &  72.9  &  56.0  \\
            UNet34~\cite{ronneberger2015u}     &  48.9  &  55.3  &  62.6  &  26.6  &  50.3  &  23.5  &  37.9  &  52.4  &  49.8  &  41.6  &  84.5  &  28.6  &  32.1  &  54.0  &  12.8  &  60.8  &  55.3  &  38.5  &  81.6  &  73.6  &  56.6  \\
            3DMV~\cite{dai20183dmv}            &  49.8  &  48.1  &  61.2  &  57.9  &  45.6  &  34.3  &  38.4  &  62.3  &  52.5  &  38.1  &  84.5  &  25.4  &  26.4  &  55.7  &  18.2  &  58.1  &  59.8  &  42.9  &  76.0  &  66.1  &  44.6  \\
            FuseNet$^\dagger$~\cite{hazirbas2016fusenet}&  53.5  &  57.0  &  68.1  &  18.2  &  51.2  &  29.0  &  43.1  &  65.9  &  50.4  &  49.5  &  90.3  &  30.8  &  42.8  &  52.3  &  36.5  &  67.6  &  62.1  &  47.0  &  76.2  &  77.9  &  54.1  \\
            SSMA$^\dagger$~\cite{valada2019self}        &  57.7  &  69.5  &  71.6  &  43.9  &  56.3  &  31.4  &  44.4  &  71.9  &  55.1  &  50.3  &  88.7  &  34.6  &  34.8  &  60.3  &  35.3  & {70.9} &  60.0  &  45.7  & {90.1} &  78.6  &  59.9  \\
            RFBNet$^\dagger$~\cite{deng2019rfbnet}       &  59.2  &  61.6  &  75.8  &  65.9  &  58.1  &  33.0  & {46.9} &  65.5  &  54.3  &  52.4  &  92.4  &  35.5  &  33.6  &  57.2  &  47.9  &  67.1  &  64.8  &  48.0  &  81.4  &  81.4  &  61.4  \\
            BPNet (Ours)$^\dagger$                     & {67.0} & {82.2} & {79.5} & {83.6} & {65.9} & {48.1} &  45.1  & {76.9} & {65.6} & {56.7} & {93.1} & {39.5} & {39.0} & {70.0} & {53.4} &  68.9  & {77.0} & {57.4} &  86.5  & {83.1} & {67.5} \\
            \bottomrule[1pt]
    \end{tabular}}
	\vspace{0.5mm}
	\caption{Comparison with the top methods on ScanNetV2 2D Semantic label benchmark, including 2D-only, 3D predictions projection, and joint 2D-3D-input based methods (marked with $\dagger$).}
    \vspace{-4mm}
    \label{tab:seg_2d}
\end{table*}


\subsection{Dataset and Metric}
\vspace{-2mm}
ScanNetV2~\cite{dai2017scannet} contains indoor scenes like offices and living rooms, with 2.5 million RGB-D frames
in more than 1500 scans, annotated with 3D camera poses, surface reconstructions, and semantic segmentation.
It is officially split into 1201 training and 312 validation scans that were taken from 706 different scenes, which means each scene was captured around one to three times, and 100 scans test set with hidden ground truth, used for the benchmark.
For the 3D input, as done in~\cite{choy20194d}, we extract points cloud from the reconstructed surfaces, and voxelize them into 3D volumes, each voxel is associate with a 3D feature ( R, G, B), while for 2D input, we use the RGB images.
Following MinkowskiNet~\cite{choy20194d}, we set the voxel size to 2cm for the benchmark results, and set it to 5cm in the ablation study for efficient training.
For the evaluation metrics, we use the mean of class-wise intersection over union (mIoU).

\subsection{Comparison on the ScanNet Benchmark}
\vspace{-1mm}
\paragraph{3D semantic segmentation.}
To evaluate the effectiveness of our BPNet for 3D semantic segmentation, we compare our method with typical streams of methods on the test set of ScnaNetV2 in Table~\ref{tab:seg_3d}, including point-based methods: PointNet++~\cite{qi2017pointnet++}, MCCNN~\cite{hermosilla2018monte}, PointASNL~\cite{yan2020pointasnl} and KP-FCNN~\cite{thomas2019kpconv}; sparse convolution based method: MinkowskiNet~\cite{choy20194d}; and joint 2D-3D-input based methods: 3DMV~\cite{dai20183dmv}, MVPNet~\cite{jaritz2019multi} and SPLATNet~\cite{su2018splatnet}.
We can see that BPNet outperforms the point-based methods, \eg, PointNet++, MCCNN, and KP-FCNN, by a large margin ($\geq 6.5$ mIoU), due to their limited receptive field and less effectiveness on feature extraction.
For the strong sparse convolution based method, MinkowskiNet, although its network receptive field is very large and global context is rich, our method also outperforms it by $1.3$ mIoU, since our method leverages the extra 2D features' information.
And importantly, our method significantly outperforms other joint 2D-3D-input based methods ($\geq 10.5$ mIoU), since their methods adopt the unidirectional scheme to combine the 2D and 3D information, that means they only regard the 2D CNN as a pre-feature-extractor without bidirectionally interacting features between 2D and 3D CNNs.
Besides, other information may further be utilized to boost the performance, like the instance annotation~\cite{han2020occuseg}.

To qualitatively evaluate the 3D semantic segmentation results, we compare our results with the strong 3D-only method, MinkowskiNet~\cite{choy20194d}, and also the ground truth in ``3D'' rows part in Figure~\ref{fig:seg_3d}.
As shown in the red boxes, MinkowskiNet cannot correctly distinguish the boundaries in ``backpack/floor'' and ``picture/wall'', while our BPNet performs well.
This is because the textures of input 3D data are too coarse, that makes it challenging to correctly predict semantics only depend on the 3D input, while our method can additionally leverage the advantages in 2D data to integrate the high-quality texture in images for better predicting semantics.
It is worth noting that, obtaining corresponding 2D input is not expensive as cameras are usually equipped with the 3D sensor in the common applications, \eg, indoor/outdoor robotics and autonomous driving.

\vspace{-4mm}
\paragraph{2D semantic segmentation.}
We then evaluate the effectiveness of BPNet for 2D semantic segmentation.
We compare BPNet with top methods on the ScanNet benchmark, including 2D-only methods: PSPNet~\cite{zhao2017pyramid} and UNet~\cite{ronneberger2015u}; method of projecting 3D predictions to 2D images: 3DMV~\cite{dai20183dmv}; and joint 2D-3D-input based methods: FuseNet~\cite{hazirbas2016fusenet}, SSMA~\cite{valada2019self} and RFBNet~\cite{deng2019rfbnet}, as shown in Table~\ref{tab:seg_2d}.
Note that, the 3D and 2D results of our BPNet in Table~\ref{tab:seg_3d} and~\ref{tab:seg_2d} are produced by the same model.
We can see that BPNet outperforms the 2D-only methods by a large margin ($\geq 18.1$ mIoU).
We admit that this comparison is not completely fair since BPNet also leverages 3D information, but such comparison gives evidence that 3D information can boost 2D semantic segmentation.
For the method of projecting 3D predictions to 2D images, like 3DMV, although the 3D information is utilized, it still cannot perform as well as BPNet as it is not tailor-made for 2D semantic segmentation.
For the joint 2D-3D-input based methods, like FuseNet, SSMA, and RFBNet, both 2D and 3D information are employed, BPNet still outperforms them a lot ($\geq 7.8$ mIoU), thanks to our bidirectional projection module (BPM) that effectively interacts with 2D and 3D features to integrate the strengths of these two visual domains.

Also, we qualitatively compare the results of 2D-only network, UNet34, BPNet, and ground truth in ``2D'' rows part in Figure~\ref{fig:seg_3d}.
We can see that BPNet not only correctly distinguishes backpack from the floor but also segments sharper and more accurate boundaries for the ``chair'' and ``table''.
It is because of the underlying geometric clues provided by 3D features.

\subsection{Ablation Study}
To explore the effectiveness of different configurations for BPNet, we conduct the following ablation experiments on the validation set of ScanNetV2.
To save training time, we use MinkowskiUNet18A for the 3D part and UNet34 for the 2D part in our BPNet, and the voxel size is set to 5cm in the first three ablations.
Since there are too many 2D frames in the validation sequences, we follow the benchmark method to sub-sample a frame from every 100 frames to form the new validation set for 2D semantic segmentation.

\vspace{-4mm}
\paragraph{Ablation for projection level.}

\begin{table}[!t]
	\centering
	\renewcommand{\tabcolsep}{7pt}

	\resizebox{1.0\linewidth}{!}{
	\begin{tabular}{l|c|cc}
		\toprule[1pt] 
		\multirow{2}{*}{Method}               &  {Projection}  &   \multicolumn{2}{c}{mIoU}    \\ \cline{3-4}
		&     Level     &      2D       &      3D       \\ \hline
		UNet34                        &     {---}      &     61.5      &     {---}     \\
		MinkowskiUNet18A                   &     {---}      &     {---}     &     68.0      \\ \hline
		Ours {W/} BPM                    &       P2       &     63.5      &     70.3      \\
		Ours {W/} BPM                    &       P3       &     64.1      &     70.5      \\
		Ours {W/} BPM                    &       P4       &     62.3      &     69.7      \\
		Ours {W/} BPM                    &       P5       &     61.8      &     68.5      \\
		Ours {W/} BPM                    & P2, P3, P4, P5 & \textbf{65.1} & \textbf{70.6} \\ \hline
		Ours {W/} UPM$_{\text{2D}\,\rightarrow\,\text{3D}}$ & P2, P3, P4, P5 &     62.2      &     69.7      \\
		Ours {W/} UPM$_{\text{2D}\,\leftarrow\,\text{3D}}$  & P2, P3, P4, P5 &     65.0      &     68.8      \\
		\bottomrule[1pt]
	\end{tabular}}
	\vspace{0.5mm}
	\caption{
		2D and 3D semantic segmentation results of different projection levels and directions on the validation set of ScanNetV2.
	}
	\label{tab:abla_level}
	\vspace{-4mm}
\end{table}

As shown in Figure~\ref{fig:method}, the BPM is applied in all the four pyramid levels (P2, P3, P4, and P5) for our full method.
For ablation, we apply BPM at a certain level and compare the results of baseline methods (UNet34 and MinkowskiUNet18A), our full method, and its variants.
From the first seven rows in Table~\ref{tab:abla_level}, we can see that our framework with BPM at every single level performs better than the 2D-only network, UNet34, or the 3D-only network,  MinkowskiUNet18A.
It means bidirectionally interacting 2D and 3D features even only at a single level can improve both 2D and 3D semantic segmentation, since useful information from the other domains is introduced.
More importantly, our framework with BPM at all the four levels outperforms all the other variants, which shows bidirectionally interacting 2D and 3D features at both low- and high-levels can better integrate the advantages in the 2D and 3D domains.

\vspace{-4mm}
\paragraph{Unidirectional projection \textit{vs.} Bidirectional projection.}
To evaluate the effectiveness of projection directions, we replace the bidirectional projection module (BPM) with the unidirectional projection module (UPM), \ie, UPM$_{\text{2D}\,\rightarrow\,\text{3D}}$ and UPM$_{\text{2D}\,\leftarrow\,\text{3D}}$, and compare the results in Table~\ref{tab:abla_level}.
We can see that our method with UPM$_{\text{2D}\,\rightarrow\,\text{3D}}$ and UPM$_{\text{2D}\,\leftarrow\,\text{3D}}$ outperforms the 2D- or 3D-only baselines, but our method with BPM performs best for both 2D and 3D results. 
It evidences that 2D and 3D semantic segmentation can better benefit each other by bidirectionally interacting features.


\begin{table}[!t]
	\centering
	\renewcommand{\tabcolsep}{6.5pt}
	\resizebox{1.0\linewidth}{!}{
	\begin{tabular}{c|ccccc}
		\toprule[1pt] 
		Number of Views &       1       &  2   &       3       &  4   &  5   \\ \hline
		2D mIoU     & \textbf{66.5} & 65.8 &     65.1      & 63.9 & 63.6 \\
		3D mIoU     &     68.1      & 68.6 & \textbf{70.6} & 70.5 & 70.5 \\
		\bottomrule[1pt]
	\end{tabular}}
		\vspace{0.5mm}
		\caption{
		2D and 3D semantic segmentation results of different view numbers on the validation set of ScanNetV2.
	}
	\label{tab:abla_view}
\end{table}

\vspace{-4mm}
\paragraph{Ablation for the number of 2D views.}

We then explore the influence of the number of 2D views, and the results are shown in Table~\ref{tab:abla_view}.
For 2D semantic segmentation, BPNet with one 2D view performs best while with five views perform worst.
This is because when there is only one view, the 3D to 2D information transformation can be best focused.
On the other hand, when the number of views increasing from one to three, the 3D semantic segmentation result becomes better and better, but it decreases slightly when the view number further increasing to five.
It may because too few views cannot provide sufficient 2D information while too many views can hinder the network from extracting useful information while discarding the redundant information.


\begin{table}[!t]
	\centering
	\renewcommand{\tabcolsep}{9pt}
	
	\resizebox{1.0\linewidth}{!}{
	\begin{tabular}{l|c|cc}
		\toprule[1pt] 
		\multirow{2}{*}{Method} & \multirow{2}{*}{Voxel Size} &   \multicolumn{2}{c}{mIoU}    \\ \cline{3-4}
		&                             &      2D       &      3D       \\ \hline
		UNet34          &            {---}            &     61.5      &     {---}     \\
		MinkowskiUNet18A     &             5cm             &     {---}     &     68.0      \\
		MinkowskiUNet18A     &             2cm             &     {---}     &     72.1      \\
		Ours           &             5cm             &     65.1      &     70.6      \\
		Ours           &             2cm             & \textbf{71.9} & \textbf{73.9} \\ 
		\bottomrule[1pt]
	\end{tabular}}
	\vspace*{0.5mm}
	\caption{
		2D and 3D semantic segmentation results of different voxel sizes on the validation set of ScanNetV2.
	}
	\label{tab:abla_voxel}
	\vspace{-5mm}
\end{table}

\vspace{-4mm}
\paragraph{Ablation for voxel size.}

Finally, we evaluate the influence of voxel size for both 3D and 2D semantic segmentation, as shown in Table~\ref{tab:abla_voxel}.
Comparing the second and third rows, we can see that more fine-grained voxels can greatly improve the 3D semantic segmentation for the 3D-only network.
And comparing the last two rows and others, we can see that decreasing voxel size can not only improve the performance of 3D semantic segmentation but also greatly improve the 2D semantic segmentation interestingly.
It evidences that the 3D information is transformed into 2D CNNs and the higher quality 3D information can better boost the 2D semantic segmentation.

\vspace{-1.5mm}
\subsection{BPNet on NYUv2}
\vspace{-1.5mm}
Although our BPNet is designed for the 3D scene with 2D images, we can also convert the RGB-D data into the 3D scene to perform semantic segmentation.
Note that, RGB-D data is known as 2.5D data rather than 3D data as the depth is view-dependent.
To evaluate the generality of BPNet on 2.5D data, we apply it to the popular RGB-D semantic segmentation dataset, NYUv2~\cite{Silberman:ECCV12}.
It contains 1,449 densely labeled pairs of aligned RGB and depth images.
We followed the official training and testing set splits, which use 795 instances for training and 654 for testing.
We converted the depth to point clouds according to the depth camera's matrix and back-projected the 2D annotations for the generated point clouds as our 3D labels.

Followed 3DMV~\cite{dai20183dmv}, we adopt the configuration of the 13-class label and report the dense pixel classification accuracy compared with typical RGB-D based methods and joint 2D-3D method 3DMV in Table~\ref{tab:nyu}.
Overall, our BPNet performs favorably against the typical RGB-D and joint 2D-3D baselines.
The results on ScanNetV2 and NYUv2 demonstrate the generality of BPNet for different types of datasets.

\begin{table}[!t]
	\centering
	\renewcommand{\tabcolsep}{12pt}
	\begin{tabular}{l|c}
		\toprule[1pt] 
		Method                             &    Accuracy      \\ \hline
		SceneNet~\cite{handa2015scenenet} & 52.5 \\
		Hermans \etal.~\cite{hermans2014dense} & 54.3 \\
		SemanticFusion~\cite{mccormac2017semanticfusion} & 59.2 \\
		Dai \etal.~\cite{dai2017scannet} & 60.7 \\
		3DMV~\cite{dai20183dmv} & 71.2 \\
		\textbf{BPNet}                     & \textbf{73.5}  \\    \bottomrule[1pt]
	\end{tabular}
	\vspace{1mm}
	\caption{Semantic segmentation results (13-class task) on NYUv2~\cite{Silberman:ECCV12} using dense pixel classification accuracy metric. Note that the reported result of Dai \etal. is on the 11-class task.}
	\vspace{-4.5mm}
	\label{tab:nyu}
\end{table}


\section{Conclusion}
\label{sec:conclusion}
In this work, we propose a bidirectional projection network, BPNet, to jointly perform 2D and 3D semantic segmentation, that leverages the complementary advantages in 2D and 3D data.
We enable the bidirectional feature interacting between 2D and 3D CNNs in multiple pyramid levels via the proposed bidirectional projection module (BPM), such that the strengths of these two visual domains can be integrated for better scene recognition.
BPNet achieves top performance on the ScanNetV2 benchmark and consistently outperforms our baseline with a single 2D/3D network. Also, with our BPNet, we can get more consistent 2D and 3D results. 
We believe the insight behind the BPM can also benefit other visual recognition tasks where 2D and 3D observations are both available, and it would advance related techniques in the community.

\paragraph{Acknowledgments.}
This project is supported by Shenzhen Science and Technology Program (No.JCYJ20180507182410327) and The Science and Technology Plan Project of Guangzhou (No.201704020141).

{\small
	\bibliographystyle{ieee_fullname}
	\bibliography{egbib}
}
\end{document}